\documentclass[10pt,letterpaper]{article}
\usepackage[T1]{fontenc}
\usepackage[utf8]{inputenc}
\usepackage[margin=0.95in]{geometry}
\usepackage{lmodern,microtype,amsmath,amssymb,graphicx,booktabs,tabularx,array,longtable}
\usepackage[authoryear,round]{natbib}
\usepackage{enumitem,placeins,xurl,flafter}
\usepackage[hidelinks,bookmarksnumbered=true]{hyperref}
\usepackage[capitalize,nameinlink,noabbrev]{cleveref}
\usepackage{listings,upquote}
\setlist{itemsep=3pt,topsep=5pt,leftmargin=*}
\newcommand{\Full}{\mathrm{F}}\newcommand{\Proj}{\mathrm{P}}\newcommand{\ind}{\mathbf{1}}

\hypersetup{pdftitle={Evaluating Budgeted Context Projection with Unexecuted Companion Runs},pdfauthor={Guangzhe Zhang},pdfsubject={Revised research preprint prepared for author review}}
\title{Evaluating Budgeted Context Projection\\with Unexecuted Companion Runs}
\author{Guangzhe Zhang\\Independent AI Researcher\\\texttt{twhite.zh@gmail.com}}
\date{October 5, 2026}
\begin{document}
\maketitle
\begin{abstract}
Context projection can shorten individual requests while changing whether an agent finishes within its budget. We examine how sequential evaluation obscures this trade-off when a capped first continuation prevents its companion from running. In a recorded ReVerPi source-reading campaign, 15 pairs with two final answers yield 12 historically scored successes per arm. Retaining all 27 intervention boundaries distinguishes observed failures from ten unexecuted companions and bounds projected-minus-full success between $-9$ and $+1$ tasks. Under the archived scoring contract, a frozen projection selector has a success difference from full context of $[-3,0]$; outside four fitting tasks, it is $[-4,-1]$ across 23 boundaries. Excluding one task whose platform premise is not established by retained actor-input evidence changes the nonfitting range to $[-3,0]$ across 22 boundaries, so strict inferiority is not robust to that exclusion. Among eleven historically joint-success pairs, projection uses 25\% fewer aggregate logical tokens but more tokens for the median pair and 55 rather than 35 suffix requests. These retrospective results concern one adaptively assembled campaign, not population performance. The case motivates accounting that retains every boundary, preserves unknown outcomes and policy dependencies, checks task premises, and separates bounded completion from success-conditioned resource use.
\end{abstract}

\section{Introduction}\label{sec:intro}
Context management changes an agent's interaction with its evidence. Replacing old tool observations with short excerpts may reduce each request, yet require additional searching, reading, and model turns before an answer is produced. Existing work therefore evaluates compression through downstream behavior and end-to-end resource use, rather than the amount of text removed alone~\citep{complexitytrap,trace,tokenreduction}. A separate question arises when the evaluation controller itself decides which continuations become observable: \emph{what can a paired comparison establish if the second continuation is never run after the first reaches its limit?}

We study this question in ReVerPi, a context-projection extension for the Pi agent harness~\citep{pi}. The evaluation uses read-only questions about Python package sources, with controller-scored JSON answers. It is not a repository-editing benchmark. At the first request eligible for projection, the controller seals a common full-context prefix and allocates full and projected continuations, each with twelve new model requests. The historical runner executes the two arms sequentially and stops the pair when an arm does not return a final answer. A wrong final answer still counts as \emph{completed} for this scheduling rule. A continuation exhausting twelve requests without an answer is \emph{capped}; a suppressed companion is \emph{unexecuted}. These are different observed states.

The distinction changes the result. The fifteen pairs in which both arms return final answers show twelve correct answers per arm. However, twelve additional runs reach the intervention boundary and then stop, including ten that leave their companion unexecuted. Treating capped arms as known failures of the stated endpoint and unexecuted arms as unknown gives a projected-minus-full success range of $-9$ to $+1$ across all 27 boundaries. A frozen rule that chooses projection on long histories has a range of $-3$ to $0$ against full context on the same frame. Under the archived scoring contract, outside its four fitting tasks even the most favorable completion leaves that rule at least one success behind full context. One task's platform premise is not established by the retained actor-input evidence. Excluding that task changes the rule's range to $[-2,+1]$ over 26 boundaries and $[-3,0]$ over 22 nonfitting boundaries, removing the strict nonfitting deficit. These are statements about fixed recorded continuations and their retained scoring contract, not estimates of future deployment performance.

The contribution is a worked, checkable account of a controller-induced observation problem in budgeted agent evaluation. We reconstruct which outcomes the scheduling rule observes or suppresses; apply established partial-identification tools while retaining dependencies between a fixed selector and its reference policy; and separate those whole-frame quality statements from resource summaries on the fully observed joint-success subset. The resulting reporting procedure specifies the boundary ledger, endpoint states, denominators, and policy relationships an evaluator must preserve. It is intended to make similar audits possible, not to introduce a new missing-data theorem, compression primitive, or validated selection algorithm.

The empirical evidence is an adaptively developed campaign using one proxy-reported model identity and one stochastic continuation per observed arm. It supplies a concrete failure mode and a reproducible calculation from extracted records. It does not establish the prevalence of that failure mode across other evaluation systems. The supplementary package supports recalculation of the reported aggregates; independently auditing all original answers or repeating the private service requires evidence that is not included. \Cref{sec:availability} distinguishes these levels of reproducibility.

\section{Related Work}\label{sec:related}
\paragraph{Agent interfaces and context management.}
SWE-agent studies how agent--computer interfaces affect software-engineering behavior~\citep{sweagent}. OpenHands provides condenser interfaces for replacing older events~\citep{openhands,openhandscond}, while MemGPT manages access across memory tiers~\citep{memgpt}. SoL-Pi combines capability-constrained harness research with archiving, recall, and efficiency mechanisms~\citep{solpi}. ReVerPi's intact history, compact request view, and addressable archive belong to this existing family; they provide the setting for the evaluation study.

\paragraph{Compression and closed-loop behavior.}
LLMLingua and Selective Context reduce prompt contents~\citep{llmlingua,selectivecontext}; Lost in the Middle shows that evidence location also matters~\citep{lostmiddle}. In interactive tasks, retained information is only one part of the problem. The Complexity Trap compares observation masking and summarization and examines trajectory length~\citep{complexitytrap}. ACON optimizes compression guidance from feedback~\citep{acon}. TRACE evaluates paired closed-loop continuations from the same environment state and studies execution instability even when information can be retrieved again~\citep{trace}. Thus neither paired continuation nor additional work after compression is new here. Our narrower question concerns continuations that the \emph{evaluation scheduler} never executes, and what remains identifiable from the resulting record.

\paragraph{Quality and resource accounting.}
AI Agents That Matter argues for evaluating accuracy and cost together~\citep{agentsmatter}. Token Reduction Is Not Cost Reduction examines cache behavior, trajectory changes, failed execution, and end-to-end cost~\citep{tokenreduction}. Robust aggregation is also important in sequential-decision evaluation~\citep{agarwal2021}. We use these distinctions rather than claiming token count as a monetary proxy. The additional issue is that outcome-dependent execution changes which quality comparisons exist before their resource costs are summarized.

\paragraph{Incomplete outcomes.}
The binary bounds below are standard worst-case partial-identification bounds~\citep{manski1990,manski2003}. A known observation mechanism does not identify an outcome in a region where it is never observed~\citep{rubin1976,petersen2012}. We also distinguish the observed jointly successful continuations from a population always-success stratum under repeated stochastic sampling, in the spirit of principal stratification~\citep{frangakis2002,zhang2003}. The application here is finite-record accounting: no missing-at-random extrapolation, randomized assignment, or population noninferiority claim is required for the reported ranges.

\section{Tasks and the Recorded Execution Protocol}\label{sec:protocol}
\subsection{A read-only source question and its evidence}
Each investigator-authored question concerns a fixed subset of Python package sources. The actor can read allowlisted files and access archived observations; it cannot execute a shell, write files, or use network tools through this adapter. It returns one JSON object with specified keys and typed values. Reference answers stay with the controller.

For example, \texttt{pathspec-util} asks about the recorded pathspec~1.1.1 snapshot. Its program creates patterns \texttt{["docs/*", "/root.txt"]}, applies them to four paths, and normalizes a path string:
\begin{lstlisting}[language=Python]
other = PathSpec.from_lines("gitwildmatch", ["docs/*", "/root.txt"])
a = other.match_file("docs/index.md")
e = normalize_file("dir\\file.py")
\end{lstlisting}
This is an abbreviated illustration, not the entire prompt. The original question also requests the package's major version; \cref{app:example} gives the task-specific question and reference answer, and distinguishes them from the adapter-rendered actor prompt. This concrete contract matters: success means a correct source answer under the stated scorer, not completion of an unrestricted coding task.

\subsection{Projection and exact archive access}
ReVerPi retains the intact history and changes only the outgoing request view. A tool result becomes eligible after appearing in two completed full-observation requests, provided that it is at least 10~KiB, is not the most recent result, and is neither an error nor an archive-recovery or revalidation result. Eligibility records transmission, not comprehension.

The projected view replaces eligible observations with a content hash, retrieval instructions, and a 1,024-byte combined head/tail excerpt. Metadata is additional to that excerpt allowance. The original bytes remain in the archive. \texttt{search\_evidence} returns literal matches with original-text excerpts; \texttt{recover\_evidence} returns an exact character interval. Both operations consume a shared allowance of three successful calls. Search can itself provide usable content, so successful recovery does not necessarily require a subsequent exact-read call. Revalidation is a separate framework capability and is not used in this source-reading campaign.

\begin{figure}[htbp]
\centering\includegraphics[width=\textwidth]{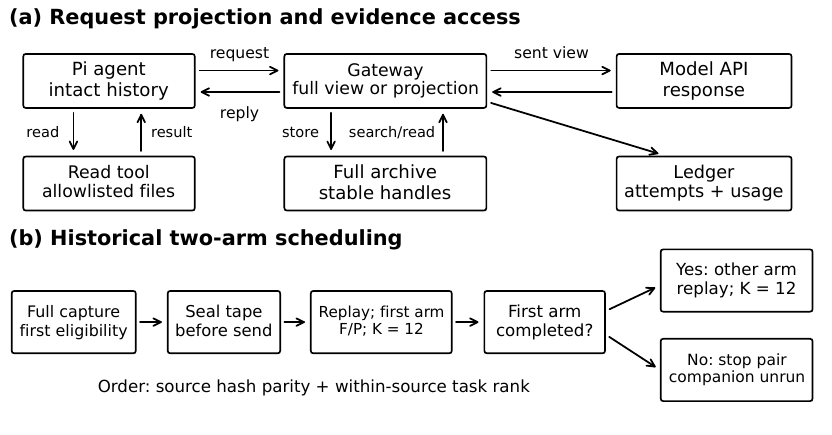}
\caption{The evaluated request view and historical scheduler. The gateway changes eligible tool observations while retaining exact archived content. The controller runs a companion continuation only if the first arm returns a final answer; a cap suppresses the companion.}
\label{fig:system}
\end{figure}

\begin{table}[tbp]
\caption{Recorded common settings. Limits are controller-side settings, not independently verified supplier limits. Additional token-admission budgets and task revisions varied across the development campaign; the supplied aggregate record does not map every variant to an immutable run configuration.}
\label{tab:settings}\centering\small
\begin{tabularx}{\textwidth}{@{}lX@{}}\toprule
Setting & Recorded value\\\midrule
Pi / protocol & 0.84.2 / Responses\\
Proxy-returned model & \texttt{deepseek-flash}\\
Effort / concurrency & low / 1\\
Capture / suffix caps & 12 / 12 newly dispatched model requests\\
Per-phase wall limit & 3,600 seconds\\
Context / output reservation & 131,072 / 65,536 tokens\\
Projection & $\geq$10 KiB; two full exposures; recency and error protection\\
Archive allowance & Three successful search/read operations in total\\
Search result & At most five matches, each with bounded excerpts\\
Exact read & At most 6,000 Unicode codepoints; default 2,000\\
Budget visibility & Character limits in schema; no live remaining-request or remaining-call counter in actor input\\\bottomrule
\end{tabularx}
\end{table}

\subsection{A shared prefix and an outcome-dependent stop}
Capture uses full observations until the first eligible request is prepared, then pauses before dispatch and seals the completed request/response tape. That point is the \emph{intervention boundary}. Full (F) and projected (P) branches replay the same tape. The contract requires their first actual suffix requests to match in tools, model settings, and all other fields except the declared tool-result representation. Later decisions and tool returns may diverge. The experimental unit is one submitted run at a realized prefix, not an API request or a newly named seed.

The controller sorts sibling task identifiers within a source. A seed-and-source SHA-256 parity chooses the first order, and sibling ranks alternate it. All 76 paired/capture plans are balanced at 38 F-first and 38 P-first, but this is deterministic allocation, not independently randomized assignment. The realized boundary subset need not retain that balance.

The historical runner breaks whenever a branch's state is not \texttt{completed}. A final but wrong answer counts as completed; a request cap without an answer does not. Ten first arms reach their caps and suppress their companions. In two further runs, full completes first and projected caps second. Thus the companion has zero observation probability under this schedule after a first-arm cap. Knowing why it is absent does not supply a comparison in that region.

\FloatBarrier
\subsection{The frozen rule and the analysis frames}
\begin{samepage}
A recorded offline search using four completed fitting pairs selects
\begin{equation}
 r_i=\ind\{h_i\geq93{,}641\},\qquad h_i=\text{history bytes},
 \label{eq:rule}
\end{equation}
where $r_i=1$ selects projected and $r_i=0$ selects full. Here $h_i$ is the UTF-8 byte length of the canonical JSON serialization of the sealed boundary's \texttt{source.messages}; it is not a token count, character count, or the length of displayed excerpts. The pinned serializer is specified in \cref{app:fitting}. The fitting tasks are \texttt{click-parameters}, \texttt{packaging-compatibility}, \texttt{boltons-indexedset}, and \texttt{tomlkit-datetimes}. They remain marked in the recorded tables.
\end{samepage}

The archived fitter~\citep{reverpiselector} enumerates two constant choices and thresholds in both directions over five predecision numerical features. Its objective is the equal-source-group mean of within-group mean selected/full logical token ratios, after rejecting candidates with observed selected harm on fitting data. This is an arithmetic mean of ratios, not a ratio of token totals. The four fitting rows give 36 candidates. None is rejected by the harm filter: the fitting set contains no full-success/projected-failure pair. Two candidates tie for the minimum cost objective, and the prescribed enumeration order selects the history rule. The exact enumeration and tied rules are documented in \cref{app:fitting}. These fitting records neither validate a harm detector nor uniquely identify history length as the preferred feature. We evaluate the recorded rule without retuning it.

\begin{table}[tbp]
\caption{Which recorded runs answer which question. These nested sets describe available information, not independent experimental sample sizes. The ten single-arm integration runs are separate from the 76 paired/capture plans.}
\label{tab:frames}\centering\small
\begin{tabularx}{\textwidth}{@{}r l X@{}}\toprule
Runs & Inclusion rule & Use\\\midrule
76 & Paired/capture plans & Campaign accounting and eligibility\\
27 & Realized paired boundary & Primary bounded-success and selector ranges\\
17 & Both bounded outcomes known & Secondary observed-expenditure policy comparison\\
15 & Both arms return a final answer & Historical completed-pair summary\\
11 & Both recorded answers correct & Resource use conditional on joint success\\\bottomrule
\end{tabularx}
\end{table}

All four fitting pairs belong to the 27-boundary, 17-comparable, and 15-completed frames; three belong to the eleven joint successes because the full packaging answer is wrong. Removing the four fitting pairs leaves 23 of the 27 boundaries and 13 of the 17 fully observed comparisons. Removing their entire source groups leaves 22 boundaries because \texttt{tomlkit-roundtrip} is also excluded. These are retrospective exclusions from development data, not untouched test sets.

\section{Outcome and Resource Accounting}\label{sec:estimands}
\subsection{Known failure and unknown outcome}
For arm $a\in\{\Full,\Proj\}$ at run $i$, define
\begin{equation}
 \begin{aligned}
 Y_{ia}=\ind\bigl\{&\text{historical scorer accepts a final answer}\\
 &\text{within }K_i\text{ newly dispatched suffix model requests}\bigr\},
 \quad K_i=12.
 \end{aligned}
 \label{eq:success}
\end{equation}
The suffix count excludes capture requests, replay of the common prefix, and tool calls. The recorded scorer uses the historical reference answer and extracted-JSON contract. These labels are retained even where a task premise warrants scrutiny (\cref{sec:pathspecvalidity}). A final answer rejected by that scorer or an observed cap without an answer gives $Y_{ia}=0$. An unexecuted arm is unknown. All observed post-boundary capped branches in this frame use twelve requests; every unexecuted companion uses zero. A stop earlier than the request cap would require additional endpoint handling if token, wall-time, integrity, or permission limits were responsible. Eventual completion with a larger allowance is a different quantity.

Encode known success, known failure, and unknown outcome as $[\ell_{ia},u_{ia}]=[1,1],[0,0],[0,1]$. For the finite recorded contrast $\Delta_N=N^{-1}\sum_i(Y_{i\Proj}-Y_{i\Full})$, standard worst-case bounds~\citep{manski1990,manski2003} give
\begin{equation}
 \Delta_N\in\left[
 \frac{1}{N}\sum_i(\ell_{i\Proj}-u_{i\Full}),
 \frac{1}{N}\sum_i(u_{i\Proj}-\ell_{i\Full})
 \right].
 \label{eq:bounds}
\end{equation}
The endpoints are attainable without cross-run restrictions on the missing binary values. This is incomplete information about fixed recorded continuations, not a confidence interval for a sampled population.

\subsection{Preserve the selector's dependence on its reference}
For a fixed rule $r_i$, its selected outcome and difference from the matched full continuation are
\begin{equation}
 Y_i^r=(1-r_i)Y_{i\Full}+r_iY_{i\Proj},
 \qquad
 Y_i^r-Y_{i\Full}=r_i(Y_{i\Proj}-Y_{i\Full}).
 \label{eq:selectorcontrast}
\end{equation}
When $r_i=0$, the difference is exactly zero, even if that full arm is unexecuted. Bounding the two policies' marginal success counts separately and subtracting would discard this shared-outcome dependence. Instead,
\begin{equation}
 \Delta_N^r\in\left[
 \frac{1}{N}\sum_i r_i(\ell_{i\Proj}-u_{i\Full}),
 \frac{1}{N}\sum_i r_i(u_{i\Proj}-\ell_{i\Full})
 \right].
 \label{eq:selectorbounds}
\end{equation}
This is a direct application of the same binary bounds to the fixed policy, not a new general bounding algorithm. \Cref{app:proofs} gives the rowwise construction and its sharpness.

\subsection{Resource use on a fully observed success subset}
Let $J=\{i:Y_{i\Full}=Y_{i\Proj}=1\}$. Each of the twelve incomplete pairs has an observed zero, so no missing companion can enter $J$. Exactly eleven recorded pairs belong to this subset. Its membership is identified even though other outcomes remain unknown. Comparing their costs describes these realized jointly successful continuations; it does not identify a population always-success effect under new stochastic draws~\citep{frangakis2002,zhang2003}.

Let $B_i$ be shared-prefix tokens, $q_{ia}$ newly dispatched suffix requests, and $\bar\tau_{ia}$ mean tokens per suffix request. Then
\begin{equation}
 S_{ia}=q_{ia}\bar\tau_{ia},\qquad
 T_{ia}=B_i+S_{ia}.
 \label{eq:cost}
\end{equation}
$T_{ia}$ is the logical expenditure of one arm including its prefix once. The acquisition expenditure for the paired experiment is $B_i+S_{i\Full}+S_{i\Proj}$, rather than two copies of the acquired prefix. Failed arms' token totals remain observed expenditures, not cost to a correct answer.

For $R_i=T_{i\Proj}/T_{i\Full}$ on $J$, the pooled token ratio weights tasks by full-arm cost:
\begin{equation}
 \frac{\sum_{i\in J}T_{i\Proj}}{\sum_{i\in J}T_{i\Full}}
 =\sum_{i\in J}w_iR_i,\qquad
 w_i=\frac{T_{i\Full}}{\sum_{j\in J}T_{j\Full}}.
 \label{eq:weighted}
\end{equation}
The median describes the middle pair; the geometric mean describes symmetric multiplicative change. These summaries need not agree. The fitting objective in \cref{sec:protocol} uses a further, explicitly different weighting scheme.

Reported tokens are uncached input $U$, cached input $C$, and output $O$, with $C$ already included in total input. Monetary cost under independently specified per-million prices is $(p_uU+p_cC+p_oO)/10^6$. The private bill and price schedule are unavailable, so no dollar-saving claim is made.

\section{Results}\label{sec:results}
\subsection{The complete intervention frame}
The campaign contains 86 runs and 641 completed model requests, with 7,971,276 input and 231,556 output tokens. Of input tokens, 5,974,272 (74.9\%) are reported cached. These are acquisition totals, not the sample size of an intervention comparison. Every response carries the proxy-returned label \texttt{deepseek-flash}; the label does not authenticate backend weights.

\begin{figure}[!ht]
\centering\includegraphics[width=\textwidth]{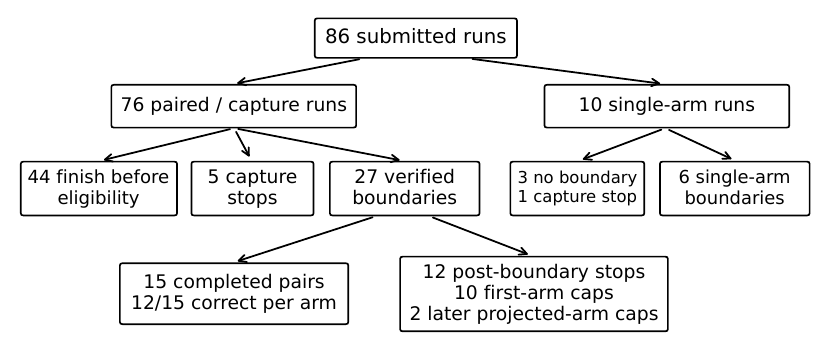}
\caption{Accounting for all 86 recorded runs. The primary paired frame has 27 boundaries: fifteen pairs with two final answers and twelve stopped pairs. Ten first-arm caps suppress a companion; two later projected caps follow full completion. Single-arm integration runs do not fill those missing companions.}
\label{fig:flow}
\end{figure}

Of the 76 paired/capture plans, 44 finish before eligibility, five stop during capture, and 27 reach a verified intervention boundary (\cref{fig:flow}). The ten single-arm integration runs contribute six other boundaries and are reported separately in \cref{tab:single}. A long trajectory or a prepared operation alone is not evidence that intervention occurred.

\subsection{The completed-pair tie does not determine whole-frame quality}
The fifteen completed pairs include eleven jointly correct, two jointly incorrect, one full-only correct, and one projected-only correct result. Each arm is correct in 12/15. Retaining the twelve stopped boundary runs instead gives \cref{tab:outcomes}.
\begin{table}[!ht]
\caption{Outcome states over all 27 paired boundaries. A cap without a correct final answer is a known bounded failure; an unexecuted arm remains unknown.}
\label{tab:outcomes}\centering
\begin{tabular}{lrr}\toprule
Recorded endpoint & Full & Projected\\\midrule
Correct & 14 & 12\\
Known bounded failure & 6 & 12\\
Unexecuted & 7 & 3\\\bottomrule
\end{tabular}
\end{table}

Thus full has 14--21 possible successes and projected has 12--15. \Cref{eq:bounds} yields $\Delta_{27}\in[-9/27,1/27]$, or $[-33.3,+3.7]$ percentage points. Even the missing outcomes most favorable to projection allow at most one extra success. The interval also permits substantial harm; it does not establish superiority or inferiority in a future population.

All first arms are observed. There are fifteen F-first boundaries, with nine correct answers, three wrong answers, and three caps, versus twelve P-first boundaries, with five correct answers and seven caps. These first-period counts avoid companion missingness but compare different adaptively obtained cases under deterministic order, not a randomized treatment effect.

\subsection{A full-frame comparison of the frozen selector}
The frozen rule selects projected in eight of the 27 boundary runs: five fully observed bounded-outcome comparisons and three projected-capped/full-unexecuted cases. When the rule selects full, the policy comparison retains exactly the same reference outcome. Applying \cref{eq:selectorbounds} produces \cref{tab:selectorbounds}.

\begin{table}[tbp]
\caption{Sharp frozen-selector-minus-full success ranges on fixed recorded frames. The count range is divided by the stated number of boundaries to obtain the percentage-point range. Exclusions are retrospective development analyses.}
\label{tab:selectorbounds}\centering\small
\begin{tabularx}{\textwidth}{@{}Xrrr@{}}\toprule
Frame & Runs & Success-count range & Difference (pp)\\\midrule
All paired boundaries & 27 & $[-3,0]$ & $[-11.11,0]$\\
Outside four fitting tasks & 23 & $[-4,-1]$ & $[-17.39,-4.35]$\\
Outside four fitting sources & 22 & $[-4,-1]$ & $[-18.18,-4.55]$\\\bottomrule
\end{tabularx}
\end{table}

The construction is transparent. The fitting case \texttt{packaging-compatibility} contributes one selector success over full. The nonfitting \texttt{pathspec-util} contributes one loss. Three projected-first capped cases---\texttt{docutils-tree}, \texttt{markdown-extensions}, and \texttt{tabulate-formats}---each contribute between $-1$ and $0$ because their full outcomes are unknown. All other policy differences are zero. Hence the total is $[-3,0]$. Removing the four fitting tasks removes the packaging gain and gives $[-4,-1]$. Removing their source groups also removes \texttt{tomlkit-roundtrip}, on which the rule chooses full and therefore has zero difference.

Under the archived scoring contract, the frozen selector cannot exceed full context's success count in this record, and outside its fitting tasks it is behind by at least one success. This follows from the fixed rule and recorded endpoints; it is not a prospective risk guarantee or evidence about an independently sampled population. Strict-format scoring leaves these policy contrasts unchanged (\cref{app:selector}). The strict nonfitting deficit depends on retaining \texttt{pathspec-util}, whose platform premise requires the following sensitivity analysis.

\subsection{Sensitivity to an unresolved task premise}\label{sec:pathspecvalidity}
The frozen pathspec source makes \texttt{normalize\_file}'s default separators depend on \texttt{os.sep} and \texttt{os.altsep}. For the string in \cref{app:example}, the recorded reference preserves the backslash, as under POSIX; under Windows the default normalization produces a slash. The retained task question and adapter-rendered prompt specify the package snapshot but not the platform. The original transmitted system prompt has not been recovered in the supplied evidence, so we cannot establish whether other actor-visible context resolved this premise. This does not prove the archived reference answer is wrong, and we do not rescore the recorded answers.

Instead, \cref{tab:pathspecsensitivity} removes the entire \texttt{pathspec-util} pair from each fixed frame. Its known selector-minus-full contribution is $-1$, so removing it adds one to both count endpoints. The nonfitting upper endpoint becomes zero: strict inferiority is no longer established, although a positive selector advantage remains excluded on that reduced nonfitting frame. Across all reduced boundaries, even that weak ordering is no longer fixed. This is a task-exclusion sensitivity, not a repaired historical experiment or a newly selected rule.
\begin{table}[htbp]
\caption{Selector-minus-full sensitivity excluding \texttt{pathspec-util}. Historical scoring is retained on every included row. Strict-format scoring gives the same contrast ranges.}
\label{tab:pathspecsensitivity}\centering\small
\begin{tabularx}{\textwidth}{@{}Xrrr@{}}\toprule
Reduced frame & Runs & Count range & Difference (pp)\\\midrule
All boundaries except pathspec-util & 26 & $[-2,+1]$ & $[-7.69,+3.85]$\\
Outside fitting tasks, also excluding pathspec-util & 22 & $[-3,0]$ & $[-13.64,0]$\\
Outside fitting sources, also excluding pathspec-util & 21 & $[-3,0]$ & $[-14.29,0]$\\\bottomrule
\end{tabularx}
\end{table}

A separate sensitivity retains every boundary but relaxes the validity of this task's accepted full score to $[0,1]$, while its observed projected cap remains zero. This is an alternative endpoint-validity assumption, not a change to the archived score and not a claim that full was unexecuted. The selector count ranges become $[-3,+1]$ over all 27 boundaries, $[-4,0]$ over 23 nonfitting-task boundaries, and $[-4,0]$ over 22 nonfitting-source boundaries. Both sensitivities therefore remove the strict nonfitting deficit.

The joint-success resource subset is unchanged by this exclusion because \texttt{pathspec-util} is not jointly successful. Its removal does affect observed policy expenditure, as reported next.

\subsection{Observed policy expenditure and fitting participation}
Seventeen boundaries have both bounded outcomes known: the fifteen completed pairs plus two full-complete/projected-capped pairs. \Cref{tab:fit} compares observed expenditure on exactly these cases, including failed trajectories. It cannot supply costs for the ten absent companions.
\begin{table}[tbp]
\caption{Frozen selector versus full on the seventeen fully observed bounded-outcome comparisons. Tokens include each logical arm's shared prefix once. Outside fitting denotes development data, not an untouched test set.}
\label{tab:fit}\centering\small
\begin{tabular}{@{}lrrrrr@{}}\toprule
Frame & Runs & F failures & Rule failures & F tokens & Rule tokens\\\midrule
Fitting & 4 & 1 & 0 & 404,703 & 298,261\\
Outside fitting & 13 & 2 & 3 & 1,419,197 & 1,541,879\\
All comparable & 17 & 3 & 3 & 1,823,900 & 1,840,140\\\bottomrule
\end{tabular}
\end{table}

\begin{samepage}
The overall failure tie includes a fitting-data benefit: packaging removes one full failure, whereas the selected pathspec cap adds one outside fitting. The thirteen outside-fitting comparisons spend 122,682 more logical tokens under the rule, an increase of 8.6\%. Across all seventeen, the increase is 16,240 tokens (0.9\%). These resource differences remain specific to the observed-comparison subset; they are not costs over all 23 nonfitting boundaries. Excluding \texttt{pathspec-util} leaves twelve comparable nonfitting cases with two failures per policy: full uses 1,295,456 logical tokens and the selector 1,264,711, a reduction of 30,745. Thus the observed nonfitting excess expenditure, as well as the strict success deficit, depends on retaining that task. This reduced comparison remains a conditional summary of observed development trajectories.\par
\end{samepage}

The historical fitting artifact's cost ratio of 0.8473 is the arithmetic mean of the four selected/full task ratios, with equal source weights. The fitting row in \cref{tab:fit} instead has ratio of totals $298{,}261/404{,}703=0.7370$. The figures answer different weighting questions. A later hindsight threshold of 95,519 obtains two failures on the seventeen cases, but it is another fit on observed development data. Its definition in \cref{app:hindsight} is separate from the original fitting procedure in \cref{app:fitting}.

\subsection{Joint success reduces aggregate tokens but adds interaction}
\begin{figure}[!ht]
\centering\includegraphics[width=\textwidth]{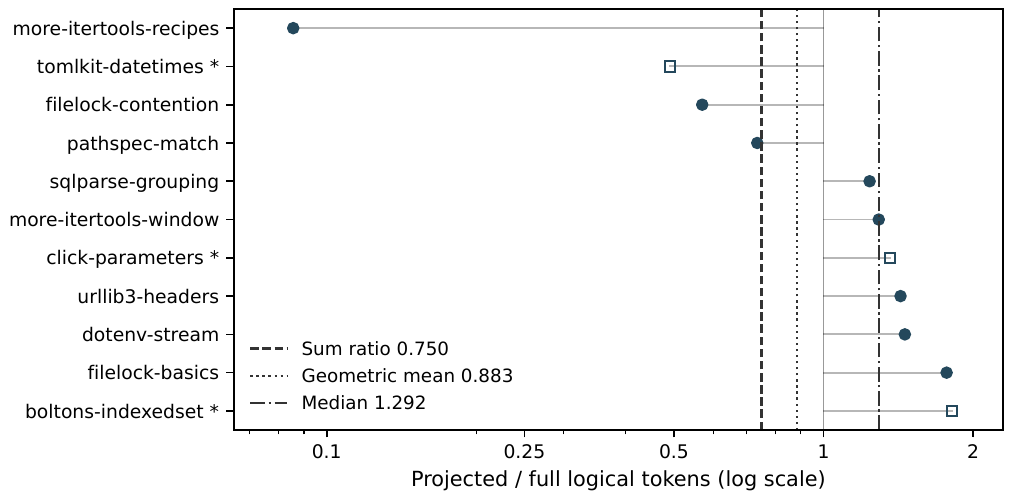}
\caption{Logical token ratios for the eleven recorded joint successes. Hollow squares and asterisks mark fitting pairs. The sum ratio, geometric mean, and median describe distinct features of a heterogeneous distribution.}
\label{fig:ratios}
\end{figure}

Within the eleven known joint successes, full uses 1,267,036 logical tokens and projected uses 949,774: a sum ratio of 0.750. The median ratio is 1.292, the geometric mean is 0.883, and seven of eleven pairs increase. \Cref{fig:ratios} shows every pair. The aggregate saving is sensitive to \texttt{more-itertools-recipes}: removing that single largest-saving case changes the sum ratio to 1.077. It remains in the primary analysis.

That case saves 382,538 tokens, exceeding the subset's net saving of 317,262; the other ten pairs jointly add 65,276 tokens. Moreover, its history is 56,382 bytes, below the selector threshold, so the frozen rule chooses full and does not obtain this observed projection saving. The 25\% sum reduction concerns projected versus full continuations on joint successes, not the frozen selector's resource benefit.

Projected uses more suffix requests in eight pairs, fewer in two, and the same number in one. Totals rise from 35 to 55, a 57\% increase. Thus fewer tokens per request and fewer requests are distinct resource objectives. Strict-format scoring removes one pair from joint success but preserves the qualitative pattern: 10 pairs have a sum ratio of 0.711, median ratio of 1.265, and 32 versus 47 requests (\cref{app:selector}).

\begin{table}[tbp]
\caption{Reported token categories for the eleven joint successes, including the shared prefix once per logical arm. Cached input is already part of total input. Complete pair-level categories are in \cref{tab:cacheall}.}
\label{tab:cache}\centering\small
\begin{tabular}{lrr}\toprule
Category & Full & Projected\\\midrule
Uncached input & 264,132 & 353,816\\
Cached input & 982,272 & 574,592\\
Output & 20,632 & 21,366\\
Total & 1,267,036 & 949,774\\\bottomrule
\end{tabular}
\end{table}

\begin{samepage}
Projection accompanies 89,684 more uncached-input, 407,680 fewer cached-input, and 734 more output tokens. Prefix rewriting and subsequent trajectory changes can both affect these categories. The record does not isolate a causal effect on a private cache. The modeled monetary difference is $(89{,}684p_u-407{,}680p_c+734p_o)/10^6$; its sign depends on prices.\par
\end{samepage}

\begin{samepage}
\subsection{Successful archive access can precede a cap}
For \texttt{pathspec-util}, history size is 94,493 bytes, so the frozen rule chooses projected. Full returns an answer accepted by the historical scorer after three suffix requests and 84,202 suffix tokens. Projected spends twelve requests and 237,629 suffix tokens without a final answer. Mean suffix-request size is about 28.1k versus 19.8k tokens: 29\% less per request but four times as many requests. Including the common 39,539-token prefix gives 123,741 versus 277,168 logical tokens.\par
\end{samepage}

After nine common-prefix reads, full makes five further reads. Projected makes thirteen reads, two searches, and three exact-read requests. The first exact read and both searches return archive text. The next two exact reads receive shared-quota refusals. Those results have \texttt{isError=false} because the extension returned an error property while the pinned Pi loop marked a tool failure only when \texttt{execute} threw~\citep{pierrorabi}. The model still saw \texttt{ReVer halted: recovery\_quota}; the structured error flag is absent from Responses tool-output items, and these short archive results are protected from projection.

\begin{figure}[!ht]
\centering\includegraphics[width=0.82\textwidth]{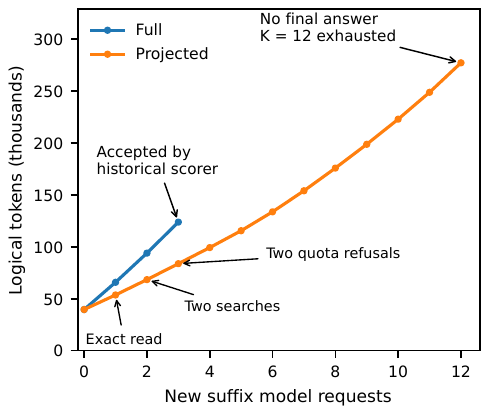}
\caption{Recorded \texttt{pathspec-util} expenditure. Acceptance denotes the historical scorer, with the platform-premise sensitivity in \cref{sec:pathspecvalidity}. Successful archive access is followed by two quota refusals and continued model requests up to the cap. Tool events are annotated at the model response requesting them; tool calls are not counted as additional model requests.}
\label{fig:pathspec}
\end{figure}

\FloatBarrier
The mismatch was corrected after the campaign~\citep{reverpierrorfix}. That change does not alter the archived trajectories or demonstrate improved completion. This example establishes successful retrieval followed by quota exhaustion and a cap. It does not establish that the retrieved fragments were sufficient, that the quota alone caused failure, or that a corrected flag would have changed the answer.

\FloatBarrier
\section{A Reporting Procedure and Its Limits}\label{sec:discussion}
The case suggests a concrete procedure for paired, budgeted agent evaluations. Its components are established accounting principles; their value here is making the observation process and the resulting policy comparison explicit.
\begin{enumerate}
\item \textbf{Declare the intervention frame.} Assign a run identifier at every valid boundary and retain it regardless of either arm's later endpoint. Separately count runs that never become eligible.
\item \textbf{Allocate both continuations before observing either.} Reserve resources per arm and decide execution independently of the first arm's ordinary cap or answer correctness. Integrity faults, unsafe environments, or revoked permission can require stopping, but must leave an explicit unresolved outcome.
\item \textbf{Record distinct endpoint states.} Preserve correct answer, incorrect final answer, request cap, other stop, and unexecuted status, together with the actual allowance and expenditure.
\item \textbf{Retain policy relationships.} Record the frozen selection decision and fitting participation. When a selector chooses its reference arm, reuse that same outcome rather than treating the policies as independent missing values.
\item \textbf{Report whole-frame quality before conditional resources.} Give known outcomes and attainable ranges on the complete frame. State the denominator of every successful-task cost summary and report requests, prefix and suffix tokens, and cache categories separately.
\item \textbf{Separate development from confirmation.} Preserve task revisions, configuration changes, tuning objectives, and all post-outcome decisions. A retuned threshold or a later implementation fix creates a new intervention to evaluate, not a repair to an old outcome.
\end{enumerate}

The present evidence shows why these records matter. The completed-pair filter suppresses relevant capped outcomes, while independent marginal ranges would lose the cancellation between a full-selecting rule and full context. The resulting policy-specific ranges can be more informative than either an incomplete-case tie or a generic missing-data warning. Testing how often such execution patterns occur in other harnesses remains future work.

\subsection{Limits of interpretation}
The task set was constructed adaptively over installed source subsets, often with two questions per source and repeated task histories. Source labels do not establish independence or representative sampling. The first-arm order is deterministic, one stochastic continuation is observed per arm, and the model identity and token usage come from a private proxy. The controlled read-only adapter does not represent the complete Pi product. Sequential execution can interact with provider or cache state. These limitations constrain intervention and population claims even when the finite-record arithmetic is exact.

The eligibility rule also selects opportunities that have already incurred repeated large observations. Forty-four of 76 paired/capture runs finish before any boundary. The campaign does not estimate deployment opportunity frequency or whether eligibility correlates with greater remaining difficulty. Forcing extra reads to create eligibility would change the question.

The pathspec question also contains a platform-dependent normalization field whose premise is not established by the retained actor-input evidence. The task-exclusion analysis in \cref{sec:pathspecvalidity} shows that the strict nonfitting success deficit and excess expenditure depend on retaining this pair. This evidence gap must not be resolved by silently adding a platform statement to the historical prompt.

The pathspec trace does not isolate a unique mechanism. Projection, retrieval allocation, absent live budget signals, error handling, and stochastic behavior remain possible contributors. Correcting the runtime flag alone has not been shown to improve task completion. A confirmatory study should freeze tasks, model identity, scorer, configuration, and selector; execute both arms independently; and distinguish repeated stochastic variation from between-task variation.

A focused methodological test would acquire complete pairs first, then apply alternative stopping and filtering rules offline to the same complete dataset. This would compare the exact full-data contrasts with complete-case summaries and missing-outcome bounds without confusing the reporting rule with a new set of tasks. Withholding outcome records offline isolates information loss caused by the reporting rule; it does not recreate cache, calendar-time, or order effects caused by actually skipping a continuation. Repeated fixed-prefix sampling of both the largest-saving and adverse cases, followed by separate quota and budget-visibility interventions, could test stability and mechanism. These studies have not been conducted, and new continuations must not be silently substituted for the ten historical missing outcomes.

\subsection{Data and code availability}\label{sec:availability}
An earlier preprint presents the same historical campaign~\citep{reverpipreprint}. The present policy-bound and scoring-sensitivity analyses reuse that record and add no model calls. The supplementary material contains the extracted campaign census, token categories, fitting rows and rule, scoring-sensitivity fields, task and projection examples, and calculation scripts. The original calculations and figure inputs are under \nolinkurl{anc/}; \nolinkurl{anc/offline_reanalysis/} adds complete-frame policy ranges, rowwise witnesses, strict-score resource summaries, and a reconstructed candidate trace. These support recalculation of the numerical inputs to the reported tables and figures without provider calls. Historical source code is available in the ReVer\_Pi repository~\citep{reverpirepo}; the fixed selector implementation and subsequent error-handling fix have distinct revision references.

Recomputing those aggregates is different from independently auditing every endpoint or repeating the service. The complete raw requests and responses, private provider settings, session/ledger databases, invoices, and all original task/answer evidence are not distributed in this package. The available digests establish consistency of supplied bytes, not experimental chronology, backend weights, or billing. Configuration variants include token-admission budgets and task revisions; the supplied census establishes twelve-request paired suffix caps but does not reconstruct every configuration variant. \Cref{app:provenance} records the availability boundary.

\subsection{Use of artificial intelligence tools}
OpenAI Codex assisted with drafting, editing, and analysis checking in preparation of this manuscript. This assistance does not supply independent experimental evidence or replace author responsibility for the reported claims.

\section{Conclusion}
The historical completed-pair comparison reports twelve correct answers per arm, but the scheduler omits ten companions after first-arm caps. Retaining all 27 intervention boundaries bounds projected-minus-full success between $-9$ and $+1$ tasks. Under the historical scorer, a fixed selector comparison preserving shared full outcomes is sharper: $[-3,0]$ over all boundaries and $[-4,-1]$ over 23 nonfitting boundaries. Excluding one task with an unresolved platform premise changes these to $[-2,+1]$ over 26 and $[-3,0]$ over 22, so strict inferiority is not robust to that exclusion. The fully observed joint-success subset separately shows lower aggregate tokens alongside greater median expenditure and more interaction. This case supports a practical evaluation discipline: retain the intervention frame, distinguish failure from nonexecution, preserve policy dependencies, and state which outcomes and resources each summary describes.

\FloatBarrier
\bibliographystyle{plainnat}\bibliography{references}
\clearpage\appendix

\FloatBarrier
\section{Finite-Frame Bounds and Their Witnesses}\label{app:proofs}
For run $i$, the unresolved binary coordinates of $(Y_{i\Full},Y_{i\Proj})$ range over their recorded intervals. The smallest contribution to $Y_{i\Proj}-Y_{i\Full}$ is $\ell_{i\Proj}-u_{i\Full}$ and the largest is $u_{i\Proj}-\ell_{i\Full}$. In the absence of cross-run restrictions, each extreme can be assigned simultaneously. Summing proves \cref{eq:bounds} and sharpness of its endpoints. Independence is unnecessary for this fixed-record calculation. Feasible means lie on an integer-over-$N$ lattice, rather than filling every real value between the endpoints.

For the selector, multiply each contribution by fixed $r_i$. Terms with $r_i=0$ vanish identically, including terms with unknown full outcomes. Terms with $r_i=1$ use the same binary construction. This proves \cref{eq:selectorbounds}. \Cref{tab:witness} lists all eight projected selections; the remaining nineteen boundaries select full and contribute zero.

\begin{table}[htbp]
\caption{Rowwise witness for the frozen-selector-minus-full count range. An asterisk marks a fitting task. Among the eight projected selections, three have unknown full outcomes.}
\label{tab:witness}\centering\small
\begin{tabular}{@{}lccc@{}}\toprule
Task & $Y_\Full$ & $Y_\Proj$ & Selected minus full\\\midrule
\texttt{packaging-compatibility}$^*$ & 0 & 1 & $+1$\\
\texttt{tomlkit-datetimes}$^*$ & 1 & 1 & $0$\\
\texttt{filelock-contention} & 1 & 1 & $0$\\
\texttt{more-itertools-window} & 1 & 1 & $0$\\
\texttt{pathspec-util} & 1 & 0 & $-1$\\
\texttt{docutils-tree} & ? & 0 & $[-1,0]$\\
\texttt{markdown-extensions} & ? & 0 & $[-1,0]$\\
\texttt{tabulate-formats} & ? & 0 & $[-1,0]$\\\midrule
Other nineteen boundaries & \multicolumn{2}{c}{Rule chooses full} & $0$\\
Total over 27 & & & $[-3,0]$\\\bottomrule
\end{tabular}
\end{table}

Choosing full success for all three unresolved rows attains the lower endpoint; choosing full failure for all three attains the upper endpoint. Removing the four fitting tasks removes the $+1$ packaging contribution and three zero terms, yielding $[-4,-1]$ over 23 runs. Excluding all fitting sources also removes the zero term for \texttt{tomlkit-roundtrip}, yielding the same count range over 22 runs. This last frame contains seventeen source groups, but grouping does not convert it into independent population sampling.

Marginally, the selector has fourteen known successes, nine known failures, and four unknown outcomes. Full has fourteen successes, six failures, and seven unknown outcomes. Subtracting their marginal ranges gives the valid but nonsharp outer bound $[-7,+4]$. Its extreme endpoints incorrectly allow the policies' shared full outcomes to differ. Preserving those shared outcomes yields the sharp range $[-3,0]$.

For the task-exclusion sensitivity, remove the fixed $-1$ contribution of \texttt{pathspec-util}; both count endpoints increase by one and each frame loses one boundary. For the endpoint-validity sensitivity, retain the row but replace only its full-score interval $[1,1]$ with $[0,1]$. Its selector contribution becomes $[-1,0]$, increasing only the upper endpoint by one. These alternative analyses are separately labelled; they do not modify the archived historical labels.

Membership in the joint-success subset has a simpler argument. The fifteen completed pairs identify eleven $(1,1)$ outcomes and four others. Each of the twelve stopped pairs has an observed zero. Thus all sixteen nonmembers are excluded regardless of the unexecuted companions. The identified eleven-member subset is a property of these realized continuations; it is not a latent population stratum under future stochastic resampling.

The companion suppression is also an observation-positivity failure. After a first-arm cap, its companion's observation probability is zero under the implemented schedule. Inverse-observation weighting would divide by zero in that region, while regression would require extrapolation into it. The stated bounds leave that extrapolation unspecified.

\clearpage
\FloatBarrier
\section{Complete Recorded Tables}\label{app:tables}
\begin{table}[!htbp]
\caption{All fifteen completed pairs. An asterisk identifies one of the four fitting tasks. $T$ is logical prefix-plus-suffix expenditure, $q$ suffix requests, and $Y$ the historical extracted-answer score. Costs in rows with an incorrect answer are expenditure, not cost to a correct answer.}
\label{tab:complete}\centering\small\setlength{\tabcolsep}{3.5pt}

\begin{tabular}{@{}lcrccrrr@{}}
\toprule
Task & First & History bytes & $Y_F/Y_P$ & $q_F/q_P$ & $T_F$ & $T_P$ & $T_P/T_F$ \\
\midrule
click-parameters$^{*}$ & F & 62,102 & 1/1 & 3/8 & 76,016 & 103,353 & 1.360 \\
packaging-compatibility$^{*}$ & F & 95,519 & 0/1 & 1/1 & 57,171 & 51,359 & 0.898 \\
boltons-indexedset$^{*}$ & F & 52,266 & 1/1 & 2/6 & 73,859 & 133,797 & 1.812 \\
cerberus-errors & F & 52,968 & 0/0 & 2/5 & 48,494 & 63,500 & 1.309 \\
dotenv-stream & P & 25,110 & 1/1 & 2/5 & 26,662 & 38,865 & 1.458 \\
filelock-basics & F & 72,397 & 1/1 & 3/11 & 91,143 & 161,204 & 1.769 \\
filelock-contention & P & 96,013 & 1/1 & 3/4 & 114,864 & 65,465 & 0.570 \\
funcy-colls & F & 49,128 & 0/0 & 2/4 & 44,800 & 52,456 & 1.171 \\
more-itertools-recipes & P & 56,382 & 1/1 & 9/1 & 418,364 & 35,826 & 0.086 \\
more-itertools-window & F & 110,836 & 1/1 & 1/2 & 63,993 & 82,647 & 1.292 \\
pathspec-match & P & 63,576 & 1/1 & 2/2 & 66,448 & 48,875 & 0.736 \\
six-moves & F & 39,668 & 1/0 & 2/5 & 51,560 & 79,484 & 1.542 \\
sqlparse-grouping & P & 89,881 & 1/1 & 2/6 & 75,974 & 94,050 & 1.238 \\
tomlkit-datetimes$^{*}$ & F & 93,641 & 1/1 & 6/5 & 197,657 & 97,027 & 0.491 \\
urllib3-headers & F & 62,556 & 1/1 & 2/5 & 62,056 & 88,665 & 1.429 \\
\bottomrule
\end{tabular}

\end{table}

\begin{table}[htbp]
\caption{All twelve stopped paired/capture boundary runs. A question mark denotes an unexecuted arm; an observed capped arm has $Y=0$. Rule selects projected when history bytes are at least 93,641. In ten rows the first arm caps and the companion is unexecuted; in the other two, F completes and P subsequently caps.}
\label{tab:stopped}\centering\small
\begin{tabular}{@{}lcrccc@{}}
\toprule
Task & First & History bytes & Rule & $Y_F/Y_P$ & $q_F/q_P$ \\
\midrule
attrs-define & P & 58,749 & F & ?/0 & 0/12 \\
docutils-nodes & F & 72,019 & F & 1/0 & 8/12 \\
docutils-tree & P & 154,876 & P & ?/0 & 0/12 \\
markdown-extensions & P & 130,201 & P & ?/0 & 0/12 \\
marshmallow-fields & P & 51,348 & F & ?/0 & 0/12 \\
pathspec-util & F & 94,493 & P & 1/0 & 3/12 \\
pyparsing-helpers & P & 81,100 & F & ?/0 & 0/12 \\
pyparsing-parse & F & 45,227 & F & 0/? & 12/0 \\
rich-table & F & 62,360 & F & 0/? & 12/0 \\
sqlalchemy-core & F & 48,268 & F & 0/? & 12/0 \\
tabulate-formats & P & 124,625 & P & ?/0 & 0/12 \\
tomlkit-roundtrip & P & 64,489 & F & ?/0 & 0/12 \\
\bottomrule
\end{tabular}

\end{table}

\begin{table}[htbp]
\caption{All six single-arm boundary runs and the earlier paired boundary for the same task. History $h$ is measured in bytes. These are repeat trajectories, not paired counterfactuals for the single-arm run. Two decisions change from F to P. The single-arm \texttt{pathspec-match} ends without an answer; its expenditure is partial.}
\label{tab:single}\centering\small\setlength{\tabcolsep}{3pt}\begin{tabular}{@{}lrrcccr@{}}
\toprule
Task & Paired $h$ & Repeat $h$ & Choice & $q$ & Outcome & Tokens \\
\midrule
filelock-basics & 72,397 & 77,200 & F $\to$ F & 2 & correct & 79,050 \\
filelock-contention & 96,013 & 123,601 & P $\to$ P & 10 & correct & 206,300 \\
more-itertools-recipes & 56,382 & 110,194 & F $\to$ P & 2 & correct & 100,078 \\
pathspec-match & 63,576 & 97,045 & F $\to$ P & 12 & capped & 244,370 \\
sqlparse-grouping & 89,881 & 64,817 & F $\to$ F & 5 & correct & 135,759 \\
urllib3-headers & 62,556 & 54,818 & F $\to$ F & 2 & correct & 49,185 \\
\bottomrule
\end{tabular}
\end{table}

\begin{table}[htbp]
\caption{Complete token categories on the observed jointly correct stratum $J$. $U$ is uncached input, $C$ cached input, $O$ output, and $q$ suffix requests. Categories include capture once for each logical arm. Fitting pairs are marked with an asterisk.}
\label{tab:cacheall}\centering\small\setlength{\tabcolsep}{3.5pt}
\begin{tabular}{@{}lrrrrrrrr@{}}
\toprule
Task & $U_F$ & $C_F$ & $O_F$ & $U_P$ & $C_P$ & $O_P$ & $q_F$ & $q_P$ \\
\midrule
more-itertools-recipes & 60,479 & 352,768 & 5,117 & 15,514 & 18,944 & 1,368 & 9 & 1 \\
tomlkit-datetimes$^{*}$ & 28,063 & 167,424 & 2,170 & 42,956 & 52,480 & 1,591 & 6 & 5 \\
filelock-contention & 24,917 & 87,552 & 2,395 & 24,380 & 38,528 & 2,557 & 3 & 4 \\
pathspec-match & 22,217 & 41,472 & 2,759 & 28,724 & 17,792 & 2,359 & 2 & 2 \\
sqlparse-grouping & 24,755 & 49,152 & 2,067 & 46,517 & 45,056 & 2,477 & 2 & 6 \\
more-itertools-window & 29,340 & 33,792 & 861 & 44,430 & 36,992 & 1,225 & 1 & 2 \\
click-parameters$^{*}$ & 18,965 & 55,424 & 1,627 & 37,691 & 62,848 & 2,814 & 3 & 8 \\
urllib3-headers & 14,984 & 45,952 & 1,120 & 27,242 & 60,032 & 1,391 & 2 & 5 \\
dotenv-stream & 5,434 & 20,736 & 492 & 9,660 & 28,032 & 1,173 & 2 & 5 \\
filelock-basics & 21,755 & 68,736 & 652 & 56,096 & 102,656 & 2,452 & 3 & 11 \\
boltons-indexedset$^{*}$ & 13,223 & 59,264 & 1,372 & 20,606 & 111,232 & 1,959 & 2 & 6 \\
\bottomrule
\end{tabular}

\end{table}

\FloatBarrier

\FloatBarrier
\section{Fitting and Sensitivity Analyses}\label{app:selector}
\subsection{The recorded search and its objective}\label{app:fitting}
The historical selector implementation is pinned to the reference revision listed in \citet{reverpiselector}. The analysis code at that revision computes the history feature as
\begin{lstlisting}[language=Python]
len(canonical(boundary["source"]["messages"]).encode("utf-8"))
\end{lstlisting}
The canonical serializer uses \texttt{json.dumps} with \texttt{ensure\_ascii=False}, \texttt{sort\_keys=True}, compact comma/colon separators, and \texttt{allow\_nan=False}. It serializes the retained structured message list, rather than concatenating visible message text. The selector's candidate enumeration is:
\begin{enumerate}
\item Constant full, then constant projected.
\item Features in the order \nolinkurl{eligible_bytes}, \nolinkurl{history_bytes}, \nolinkurl{eligible_count}, \nolinkurl{prefix_requests}, \nolinkurl{last_cached_fraction}.
\item For each feature, distinct floating-point values in ascending order, with threshold directions $\geq$ then $\leq$ at each value.
\end{enumerate}
The implementation caps a feature's grid at 32 values using rounded, evenly spaced indices when there are more than 32 distinct values. The cap is inactive here: the five fitting-feature grids have sizes 4, 4, 2, 3, and 4, respectively. Thus there are $2+2(4+4+2+3+4)=36$ candidates. Full-precision fitting features are retained in the supplied fitting records.

For a candidate $r$, let $I_g$ denote the fitting records from source group $g$, and let $G$ be the number of such groups. The objective is
\begin{equation}
 L(r)=\frac{1}{G}\sum_g\frac{1}{|I_g|}
 \sum_{i\in I_g}\frac{(1-r_i)T_{i\Full}+r_iT_{i\Proj}}{T_{i\Full}}.
 \label{eq:fittingobjective}
\end{equation}
Candidates with any observed $r_iY_{i\Full}(1-Y_{i\Proj})=1$ are rejected. Among the remainder, the implementation minimizes the pair $(L(r),\text{enumeration index})$, so an earlier candidate wins an exact tie. Four separate fitting sources make $L$ the ordinary mean of four ratios in this instance. For the recorded threshold, $L=0.84730644845$, whereas the ratio of fitting token totals is $0.736987371$.

All 36 candidates pass the observed-harm filter because no fitting pair has $(Y_\Full,Y_\Proj)=(1,0)$. Thus the constraint eliminates zero candidates in this search. Two candidates attain the same minimum objective: zero-based index 14 is $\texttt{history\_bytes}\geq93{,}641$, and index 31 is $\texttt{last\_cached\_fraction}\leq0.8012967413783466$. They make the same four fitting decisions, selecting projected for packaging and tomlkit, and full for click and boltons. Enumeration order chooses index 14. The data do not establish that the history feature is uniquely better, or that the filter has demonstrated protection against harm.

The archived implementation and fitting rows permit reconstruction of the search. The historical runner writes the winning rule rather than a per-candidate score trace. A trace reconstructed during reanalysis must therefore be distinguished from an originally logged artifact. Version identity and deterministic reconstruction establish internal consistency, not that the fitting procedure was prospectively registered or that later development cases were untouched.

\subsection{Hindsight threshold comparisons}\label{app:hindsight}
\begin{table}[htbp]
\caption{Policy comparisons on the same seventeen fully observed bounded-outcome records. The common prefix totals 510,392 tokens. The hindsight threshold minimizes failures and then tokens over this observed history-threshold family; it is neither deployed nor a bound on all possible agent improvements.}
\label{tab:policies}\centering\small
\begin{tabular}{@{}lrrrr@{}}\toprule
Policy & Failures & Suffix tokens & Logical tokens & Change vs F\\\midrule
Always-full & 3 & 1,313,508 & 1,823,900 & ---\\
Always-projected & 5 & 1,240,753 & 1,751,145 & $-4.0\%$\\
Frozen threshold 93,641 & 3 & 1,329,748 & 1,840,140 & $+0.9\%$\\
Hindsight threshold 95,519 & 2 & 1,276,951 & 1,787,343 & $-2.0\%$\\\bottomrule
\end{tabular}
\end{table}

The frozen threshold equals the history of fitting task \texttt{tomlkit-datetimes}. The hindsight threshold equals fitting task \texttt{packaging-compatibility} and lies above adverse \texttt{pathspec-util} at 94,493. Fourteen leave-one-source-out development folds select 95,519 twelve times, 96,013 once, and 93,641 once. Their concatenated held-out choices have four failures, versus three for the frozen rule. Each fold holds out a whole source group and trains on a variable number of remaining records. These folds reuse the observed development frame; their output is not independent confirmation.

\subsection{Strict-format scoring}
The historical scorer extracts the first decodable JSON object from a final answer. The alternative strict contract accepts a complete JSON object, optionally inside a complete JSON fence, and rejects prose wrappers, duplicate keys, and nonfinite values before normalization. Historical scores remain the main contract. The changed labels are both \texttt{click-parameters} arms and the no-boundary \texttt{packaging-normalization} capture.

On the seventeen comparable runs, strict scoring gives four full failures, six projected failures, and four frozen-rule failures, versus three, five, and three under historical scoring. Since the rule selects full on \texttt{click-parameters}, changing both arm labels leaves that row's selector-minus-full difference zero. Every full-frame selector contrast in \cref{tab:selectorbounds} is unchanged.

Joint success changes from eleven to ten pairs. \Cref{tab:strictJ} gives the resulting resource sensitivity. Sum ratios below one, median ratios above one, and more projected requests persist under both scores; unchanged policy failure ordering alone would not have established this resource result.
\begin{table}[htbp]
\caption{Joint-success resource summaries under the historical and strict-format scores. Logical tokens include the shared prefix once per arm.}
\label{tab:strictJ}\centering\small
\begin{tabular}{@{}lrr@{}}\toprule
Summary & Historical & Strict format\\\midrule
Jointly correct pairs & 11 & 10\\
Full logical tokens & 1,267,036 & 1,191,020\\
Projected logical tokens & 949,774 & 846,421\\
Ratio of sums & 0.749603 & 0.710669\\
Geometric mean ratio & 0.883374 & 0.846091\\
Median ratio & 1.291501 & 1.264712\\
Pairs with higher projected tokens & 7 & 6\\
Full suffix requests & 35 & 32\\
Projected suffix requests & 55 & 47\\\bottomrule
\end{tabular}
\end{table}

\FloatBarrier
\section{Original Task and Recorded Projection}\label{app:example}
The complete task-specific question for the recorded \texttt{pathspec-util} task follows. It is the \texttt{prompt} field of \nolinkurl{anc/original/data/pathspec_TASK.json}, with only line wrapping changed for display. It is not the complete actor input: the adapter prepends a development-question notice, a read-only workspace notice, the 32 allowed relative source paths, and instructions on evidence use. The rendered prompt is preserved in \nolinkurl{anc/original/data/pathspec_PUBLIC.json} and, as plain text, in \nolinkurl{pathspec_actor_prompt.txt}. System instructions and tool schemas are separate from this prompt field.
\par\noindent\begin{minipage}{\linewidth}
\begin{lstlisting}
Using only the supplied pathspec 1.1.1 source, determine the results of this program. Answer every field, consulting the source where needed:

```python
from pathspec import PathSpec
from pathspec.util import normalize_file

other = PathSpec.from_lines("gitwildmatch", ["docs/*", "/root.txt"])
a = other.match_file("docs/index.md")
b = other.match_file("docs/sub/page.md")
c = other.match_file("root.txt")
d = other.match_file("sub/root.txt")
e = normalize_file("dir\\file.py")
```

Report `docs_index` (`a`), `docs_nested` (`b`), `root_txt` (`c`), `sub_root_txt` (`d`), `normalize` (`e`) and `version_major` (`__import__("pathspec").__version__.split(".")[0]`).

Return exactly one JSON object with these six keys, no prose.
\end{lstlisting}
\end{minipage}\par
The controller reference, included only to make the scoring example explicit, is:
\begin{lstlisting}
{"docs_index":true,"docs_nested":true,"normalize":"dir\\file.py",
 "root_txt":true,"sub_root_txt":false,"version_major":"1"}
\end{lstlisting}
The actor receives the rendered prompt and allowlisted source, not this reference answer. The task-specific question computes the version with \texttt{\_\_import\_\_("pathspec")}. The adapter instructions also tell the actor to avoid fabricated extra reads, permit finishing without archive tools, and require JSON null for fields it cannot determine. Neither retained prompt field explicitly states the operating platform; the normalization field therefore requires separate scrutiny, as discussed in \cref{sec:pathspecvalidity}.

The first projected request replaces a 25,551-byte observation with metadata, instructions, and head/tail excerpts. The following display abbreviates the handle and excerpts; editorial elisions are not model-visible text. The exact replacement is supplied as \texttt{actual\_observation.txt}.
\begin{lstlisting}
[reverpi-observation-v1 {"handle":"6c69f25c...5833d22ea",
 "offset_unit":"unicode_codepoints","original_bytes":25551,
 "tool":"read","total_chars":25551}]
Historical output omitted from this request, not deleted.
This is not a current-state verification.
Read exact historical text with recover_evidence(handle, start, chars).
Locate a span with search_evidence(query, chars); use returned handle and start.
Never send query to recover_evidence.
[head]
[editorial elision of original head excerpt]
[middle omitted]
[editorial elision of original tail excerpt]
[tail]
\end{lstlisting}
The handle resolves to the original observation. The 1,024-byte excerpt allowance does not include metadata and instructions. The three successful search/read operations share one quota; subsequent refusals preserve their recorded text and false error flag. The later runtime fix does not change the historical responses or scores.

\clearpage
\FloatBarrier
\section{Source Inventory and Record Availability}\label{app:provenance}
\begin{table}[htbp]
\caption{Fixed source subsets for the 76 paired/capture plans. Each source contributes two investigator-authored plans with balanced initial order. Boundary and Pair count realized boundaries and completed pairs. File counts include licenses; complete digests are supplied with the data.}
\label{tab:sources}\centering\footnotesize\renewcommand{\arraystretch}{0.95}

\begin{tabular}{@{}lrrrl@{}}
\toprule
Source/version & Files & Boundary & Pair & Snapshot SHA prefix \\
\midrule
attrs-26.1.0 & 20 & 1 & 0 & \texttt{fc4d96802c3d} \\
boltons-26.2.0 & 31 & 1 & 1 & \texttt{b80f6739e270} \\
cachetools-7.1.4 & 9 & 0 & 0 & \texttt{1fa516c2685d} \\
cerberus-1.3.8 & 7 & 1 & 1 & \texttt{f4bb235519e6} \\
click-8.1.8 & 17 & 1 & 1 & \texttt{44353e20a2fd} \\
colorama-0.4.6 & 14 & 0 & 0 & \texttt{be58df74478d} \\
dateutil-2.9.0.post0 & 19 & 0 & 0 & \texttt{a04fe4c3b097} \\
docutils-0.23 & 130 & 2 & 0 & \texttt{95cdc9b6127e} \\
filelock-4.0.3 & 24 & 2 & 2 & \texttt{6cb9581b6f10} \\
funcy-2.1 & 17 & 1 & 1 & \texttt{f0aa75f84a32} \\
humanize-4.16.0 & 8 & 0 & 0 & \texttt{8da0a5d02e67} \\
idna-3.20 & 11 & 0 & 0 & \texttt{1a074bdb5b8c} \\
jinja2-3.1.6 & 26 & 0 & 0 & \texttt{86c4faf0e95b} \\
jmespath-1.1.0 & 9 & 0 & 0 & \texttt{c21828454ba7} \\
markdown-3.10.3 & 34 & 1 & 0 & \texttt{befd520129ae} \\
marshmallow-4.3.1 & 15 & 1 & 0 & \texttt{7b6f357d354c} \\
more-itertools-11.1.0 & 4 & 2 & 2 & \texttt{c5fa441afd8a} \\
networkx-3.7 & 584 & 0 & 0 & \texttt{c4992489e474} \\
packaging-25.0 & 19 & 1 & 1 & \texttt{c8879839459f} \\
pathspec-1.1.1 & 32 & 2 & 1 & \texttt{db9ce114b658} \\
pip-26.2.1 & 405 & 0 & 0 & \texttt{07eb31e79304} \\
platformdirs-4.11.12 & 10 & 0 & 0 & \texttt{7cc2d6b8dc56} \\
pygments-2.21.0 & 344 & 0 & 0 & \texttt{6fe831a0b547} \\
pyparsing-3.3.3 & 18 & 2 & 0 & \texttt{46640a34b12e} \\
python-dotenv-1.2.3 & 9 & 1 & 1 & \texttt{d2deeda0f0ed} \\
rich-15.0.0 & 101 & 1 & 0 & \texttt{acdbd0ecaa80} \\
six-1.17.0 & 2 & 1 & 1 & \texttt{713806dd6930} \\
sortedcontainers-2.4.0 & 5 & 0 & 0 & \texttt{150744eb5936} \\
sqlalchemy-2.0.54 & 258 & 1 & 0 & \texttt{6e6789fc1543} \\
sqlparse-0.6.0 & 22 & 1 & 1 & \texttt{71bfdfa9998d} \\
tabulate-0.10.0 & 2 & 1 & 0 & \texttt{4f0bccf42996} \\
toml-0.10.2 & 6 & 0 & 0 & \texttt{102b96e7a074} \\
tomlkit-0.15.1 & 13 & 2 & 1 & \texttt{193454b37de5} \\
toolz-1.1.0 & 32 & 0 & 0 & \texttt{ec0bd7c7eba9} \\
urllib3-2.8.0 & 37 & 1 & 1 & \texttt{d7d96f512c08} \\
voluptuous-0.16.0 & 7 & 0 & 0 & \texttt{d8c89cb9ac84} \\
wheel-0.48.0 & 16 & 0 & 0 & \texttt{c6693b2738e6} \\
xmltodict-1.0.4 & 2 & 0 & 0 & \texttt{2e0f2b772f66} \\
\bottomrule
\end{tabular}

\end{table}
\FloatBarrier

The campaign archive is identified by SHA-256
\begin{center}\scriptsize\nolinkurl{2b57e3188492a6b852684e8a85a19b92db96e7442fbb825a891ae762cf22f29c}.\end{center}
Its 12,980 listed artifacts underlie the original request census. The recorded audit cross-checks request usage with ledger attempts and phase endpoints with plan limits and sealed boundaries. The source-level auditor reports 73 passes among 76 paired/capture runs; three retain allowlist-intent rejection. Of the passes, 71 match a submitted audit and two have no submitted counterpart. Interface compliance and accounting completeness are separate checks, and rejected interface cases remain in the census.

\Cref{tab:auditrejects} identifies those three runs from the retained historical audit index. All stop during capture with reason \texttt{pi\_actor} and have no sealed boundary; none enters the 27-boundary, 17-comparable, 15-completed, or 11-joint-success frame. Excluding these interface-rejected plans therefore changes none of the paired success or cost results. They remain in the campaign accounting.
\begin{table}[htbp]
\caption{Historical audit-rejection classifications. Run IDs are \texttt{cumulative\_runs/paid/} followed by the listed task. All three stop during capture and are outside the 27/17/11 analysis frames. Counts are preserved from the historical report, not a new raw-transcript audit.}
\label{tab:auditrejects}\centering\small
\begin{tabular}{@{}lrrr@{}}\toprule
Task & \shortstack{Non-allowlisted\\read intents} & \shortstack{Aborted\\results} & \shortstack{Pending\\results}\\\midrule
\texttt{platformdirs-paths} & 1 & 1 & 0\\
\texttt{sqlparse-format} & 2 & 1 & 1\\
\texttt{wheel-metadata} & 3 & 1 & 2\\\bottomrule
\end{tabular}
\end{table}
The historical report records no successful out-of-allowlist result for these runs. A rejected intent is not evidence that content outside the allowlist was returned. The recovered report does not retain the requested paths or call identifiers, and this classification does not certify a sandbox. The safe source extract, original member hashes, and census join are supplied under \nolinkurl{anc/source_evidence/}. The separate \texttt{pathspec-util} source audit passes; that interface result does not settle its task's platform premise or independently validate answer quality.

The supplied aggregate package includes branch order, history size, fitting membership, request counts, cache categories, source digests, and strict-score sensitivity. It preserves the recorded extracted-answer contract. Full raw requests, databases, provider settings, invoices, and every task/answer pair are outside the distributed package. Consequently, an outsider can check arithmetic and the provided examples but cannot independently rescore every campaign response from these files alone.

The reference implementation is pinned at \nolinkurl{7a19d72ff86655310a4b24298191ce73cbd2c44e}. The recovery-error-signaling change is \nolinkurl{59e8497a56faddff8c0d1836456b3cd3cf5a4b6a}, and its merge is \nolinkurl{d331a4801873c3957b9ca410de43c520582197e4}. These are distinct software states. The selector file has the same Git blob at all three states; this verifies version identity of that file, not experimental chronology. The historical requests are not regenerated or replaced by later source fixes.

The six single-arm boundaries in \cref{tab:single} are repeated trajectories, not additional paired outcomes. Two decisions change from full to projected because the observed histories differ. They do not supply the missing counterfactuals at earlier captured prefixes. Earlier single-prefix experiments and calibration illustrations are not pooled into this manuscript's primary comparisons.

\end{document}